\documentclass[sigconf,nonacm]{acmart}

\AtBeginDocument{%
  }

\title{Convolutional Sparse Coding via the Locally Competitive Algorithm on Loihi 2}

\author{Geoffrey Kasenbacher}
\affiliation{
  \institution{Mercedes-Benz AG}
  \streetaddress{Leibnizstraße 2}
  \city{Böblingen}
  \country{Germany}
}
\affiliation{
  \institution{Institut für Robotik und Kognitive Systeme, Universität zu Lübeck}
  \streetaddress{Ratzeburger Allee 160}
  \city{Lübeck}
  \country{Germany}
}
\email{geoffrey.kasenbacher@uni-luebeck.de}

\author{Daniel Ruepp}
\affiliation{
  \institution{Mercedes-Benz AG}
  \streetaddress{Leibnizstraße 2}
  \city{Böblingen}
  \country{Germany}
}

\author{Gerrit A. Ecke}
\affiliation{ 
  \institution{Mercedes-Benz AG}
  \streetaddress{Leibnizstraße 2}
  \city{Böblingen}
  \country{Germany}
}

\renewcommand{\shortauthors}{Kasenbacher et al.}

\begin{abstract}
Sparse coding provides a principled framework for signal representation by expressing an input as a linear combination of only a small number of basis functions. The Locally Competitive Algorithm (LCA) is particularly attractive in the context of neuromorphic computing because its dynamics, leaky integration, thresholding, and lateral inhibition map naturally to neuromorphic hardware. While prior work has studied non-convolutional LCA on Loihi~2, the convolutional setting is of particular interest because it introduces spatial structure, weight sharing, overlapping receptive fields, and scaling behavior that are more representative of practical sparse inference workloads. In this work, we present a Loihi~2 implementation of convolutional sparse coding via the LCA and evaluate it against a conventional GPU baseline on the same inference problems. The implementation follows a one-layer recurrent LCA formulation and extends it to convolutional feature maps with local inhibitory kernels derived from pairwise filter interactions. To the best of our knowledge, this is the first implementation and benchmark of convolutional LCA on Loihi~2. Our goal is not only to demonstrate feasibility, but also to clarify in which operating regimes convolutional sparse inference becomes attractive on neuromorphic hardware. The resulting study positions convolutional LCA as a useful benchmark for structured sparse inference on emerging neuromorphic systems.
\end{abstract}


\begin{document}

\maketitle

\section{Introduction} 

Sparse coding has long been studied as an efficient representation principle for natural signals, with successful applications in denoising, compression, and inverse problems. In the sparse coding model, an input is reconstructed by combining only a small subset of elements from an overcomplete dictionary. This combination of representational flexibility and sparsity makes the approach attractive both algorithmically and biologically \cite{olshausen1996emergence, olshausen2004sparse}. However, sparse inference is typically obtained through iterative optimization, and its computational cost can become substantial as the problem dimension increases.

Among sparse inference algorithms, the Locally Competitive Algorithm (LCA) occupies a special role because it is both an optimization procedure and a neural dynamical system \cite{rozell2008sparse}. The LCA evolves through local interactions between units that integrate feedforward drive, compete through lateral inhibition, and emit thresholded activations. This structure makes it particularly well suited to neuromorphic hardware, where local state, event-driven updates, and sparse communication are native computational primitives. In that sense, the LCA is not merely compatible with neuromorphic systems; it is one of the few sparse inference algorithms whose dynamics align naturally with them.

Most prior neuromorphic studies of the LCA have focused on the non-convolutional setting \cite{fair2019sparse, hong2024memristive, parpart2023implementing}. That case is valuable as a foundational benchmark, but it omits several properties that become essential in realistic spatial inference problems. In contrast, convolutional sparse coding introduces weight sharing, overlapping receptive fields, spatially extended competition, and feature maps whose size scales with the input domain \cite{teti2022lcanets, kasenbacher2025warp}. These properties affect both the algorithmic behavior and the hardware mapping. Convolution therefore changes the problem qualitatively: it is not simply a larger version of vector sparse coding, but a distinct structured inference workload.

This work studies convolutional sparse coding via the LCA on Intel's Loihi~2 neuromorphic processor \cite{intel_labs_loihi_2021}. Our interest is twofold. First, we seek to determine whether convolutional LCA can be implemented faithfully within the constraints of Loihi~2 while preserving the essential sparse coding dynamics. Second, we seek to understand how this workload behaves relative to a GPU baselines when reconstruction quality is compared alongside hardware-dependent metrics such as latency, power, and energy.

To the best of our knowledge, this is the first implementation and benchmark of convolutional LCA on Loihi~2. Rather than framing the contribution as a simple port of an existing algorithm, we view convolutional LCA as a structured sparse-inference benchmark that exposes regime-dependent trade-offs between reconstruction quality, latency, and energy. Convolution introduces spatial reuse that is absent in the non-convolutional case, while competition remains local through overlapping receptive fields rather than global across all coefficients \cite{schultz2014replicating, zeiler2010deconvolutional}. Neuromorphic implementations can exploit such local connectivity patterns directly, making the convolutional case qualitatively different from standard vector sparse coding.

The contributions of this paper are therefore threefold. First, we present a mapping of convolutional LCA to Loihi~2 with fixed-point membrane dynamics and locally connected inhibitory interactions. Second, we establish a matched benchmarking setup against a conventional GPU implementation, separating hardware-independent metrics such as reconstruction quality from hardware-dependent metrics such as latency, power, and energy. Third, we analyze the resulting trade-offs across multiple regularization settings, thereby identifying operating regimes in which convolutional sparse inference is attractive on neuromorphic hardware.

\section{Convolutional Sparse Coding and LCA Dynamics}

Sparse coding seeks a representation of an input signal \(x\) in terms of a small number of active dictionary elements. In the convolutional setting, let \(x \in \mathbb{R}^{C \times H \times W}\) denote an input with \(C\) channels and spatial dimensions \(H \times W\). The signal is approximated by a convolutional dictionary
\(\Phi \in \mathbb{R}^{M \times C \times k_H \times k_W}\), consisting of \(M\) kernels with spatial support \(k_H \times k_W\), and coefficient maps
\(a \in \mathbb{R}^{M \times H' \times W'}\), where \(H'\) and \(W'\) denote the spatial dimensions of the coefficient maps induced by the convolutional configuration. The convolutional sparse coding problem is then written as
\begin{equation}
x \approx \Phi \,\boldsymbol{\ast}^\top\, a,
\end{equation}
and the associated inference problem becomes
\begin{equation}
E(x,\Phi,a) = \left\|x - \Phi \,\boldsymbol{\ast}^\top\, a\right\|_2^2 + \lambda \|a\|_1,
\label{eq:energy}
\end{equation}
where \(\lambda > 0\) controls the sparsity level. The first term enforces reconstruction fidelity, while the second term penalizes dense activations. Here, \(\boldsymbol{\ast}\) denotes the forward convolution (or correlation, depending on implementation convention), and \(\boldsymbol{\ast}^\top\) denotes the corresponding adjoint operation used for reconstruction.

The Locally Competitive Algorithm provides a dynamical system whose equilibrium corresponds to a sparse solution of \eqref{eq:energy}. In the convolutional case, each feature map \(i\) is associated with an internal state \(u_i(t)\) and an activation \(a_i(t)\) obtained through a thresholding nonlinearity,
\begin{equation}
a_i(t) = T_\lambda\big(u_i(t)\big).
\label{eq:lca_thresh}
\end{equation}
The corresponding dynamics can be expressed as
\begin{equation}
\tau \frac{d u_i(t)}{d t}
=
- u_i(t)
+ b_i
+ \sum_j \left(W_{ij} \ast a_j(t)\right),
\label{eq:lca_cont}
\end{equation}
where \(\tau\) is a time constant, \(b_i\) is the feedforward drive induced by the input and dictionary, and \(W_{ij}\) encodes lateral interactions between coefficient maps. In our setting, the feedforward term is obtained by convoluting the input with the convolutional dictionary, while the lateral term is derived from pairwise filter interactions and enforces competition between correlated features.

A discrete-time implementation follows from a forward Euler discretization. Writing the state at iteration \(t\) as \(u_i^{(t)}\), the update takes the form
\begin{equation}
u_i^{(t+1)}
=
(1-\tau)u_i^{(t)}
+
\tau\, b_i
+
\tau \sum_j \left(W_{ij} \ast a_j^{(t)}\right),
\label{eq:lca_disc}
\end{equation}
followed by the elementwise soft-thresholding operation
\begin{equation}
a_i^{(t+1)} = T_\lambda\big(u_i^{(t+1)}\big),
\end{equation}
where
\begin{equation}
T_\lambda(u) = \operatorname{sign}(u)\,\max(|u|-\lambda,0).
\end{equation}
This representation makes clear why the LCA is well aligned with neuromorphic hardware. The state update is local, the interaction structure is sparse and spatially structured, and the threshold nonlinearity naturally suppresses unnecessary activity.

Compared with the non-convolutional formulation, the convolutional LCA introduces several properties that are central to this study. First, the same kernels are reused across spatial positions, which makes the workload structured rather than fully connected. This shared-kernel structure is also much more practical: the dictionary remains compact even for large inputs, whereas a non-convolutional formulation would require a rapidly growing number of parameters and interactions as image size increases. Second, neighboring coefficients interact through overlapping receptive fields, which creates local inhibitory kernels whose support depends on filter size and stride. Third, the number of states grows with the spatial extent of the input, so scaling behavior becomes an intrinsic part of the benchmark.

In the experiments reported later, we use this convolutional LCA both as a sparse reconstruction model and as a benchmark workload for neuromorphic inference.

\section{Mapping Convolutional LCA to Loihi~2}

Our implementation follows the one-layer LCA formulation previously used for non-convolutional Loihi~2 realizations \cite{parpart2023implementing} and extends it to the convolutional setting. In this formulation, the sparse coefficients are represented directly by a single recurrent layer of neurons. Each neuron stores an internal membrane state \(u\), and its output corresponds to the thresholded activation \(a = T_\lambda(u)\). The feedforward drive induced by the input is precomputed and injected as a fixed bias, while recurrent synaptic connections implement the lateral competition term.

Concretely, for a given input \(x\) and dictionary \(\Phi\), we compute the feedforward term
\begin{equation}
b = \Phi \ast x,
\end{equation}
with the convolution performed at the stride associated with the current experiment. This quantity is quantized and stored as the neuron bias. The lateral interaction kernel is derived from pairwise dictionary correlations, where \(W\) denotes the filter interaction tensor obtained from the spatial convolution of the dictionary with itself. To prevent a coefficient from trivially inhibiting itself, the exact self-overlap at the center position is removed. This is expressed as
\begin{equation}
W = (\Phi \ast \Phi) - \delta,
\end{equation}
where \(\delta\) is a spatial Kronecker delta centered at the zero-shift origin. This operation is the convolutional analogue of subtracting the identity matrix in standard linear LCA formulations.

The resulting neuron dynamics are therefore of the same type as in the one-layer Loihi~2 LCA model introduced by \cite{parpart2023implementing},
\begin{equation}
u^{(t+1)} = (1-\tau)u^{(t)} + \tau\,b + \tau\,\mathrm{in}^{(t)},
\end{equation}
where
\begin{equation}
\mathrm{in}^{(t)} = W \ast a^{(t)}
\end{equation}
is the recurrent input induced by lateral competition, with \(a^{(t)}\) representing the current thresholded activations. As in prior one-layer implementations, the membrane state is represented in fixed-point form and the soft-thresholding operation is realized through a custom neuron update rule. Active units emit graded outputs corresponding to the thresholded activation, so that sparse coefficient activity is translated directly into sparse event traffic on-chip \cite{intel_labs_loihi_2021}.

After the on-chip dynamics reach the prescribed iteration budget of 1000, the final membrane states are read out and converted back to sparse coefficients using the same thresholding rule. Reconstructions are then obtained by the adjoint convolution with the learned dictionary. This yields the reconstruction-quality metrics reported later.

\section{Experimental Setup}

\subsection{Dataset and Convolutional Configurations}
\label{sec:exp_dataset}

Experiments are conducted on the Set12 image benchmark \cite{zhang2017beyond} using grayscale inputs cropped or resized to a set of fixed target sizes. As established by \cite{garcia2018convolutional}, the sparse coding objective exhibits individual convexity with respect to both the sparse representations $a$ and the dictionary filters $\Phi$. Following the approach of \cite{teti2022lcanets}, we exploit this property by employing an alternating training scheme. Specifically, we compute the sparse coefficients $a$ for each input batch using LCA dynamics, and then perform a dictionary update on $\Phi$ via SGD. This convolutional dictionary learning is performed entirely offline prior to neuromorphic evaluation. Once learned, the dictionaries are transferred to the hardware and kept fixed during inference benchmarking. This ensures that the benchmark isolates sparse inference behavior on hardware rather than conflating inference with on-chip dictionary adaptation.

\begin{figure}[t]
    \centering
    \includegraphics[width=0.42\columnwidth]{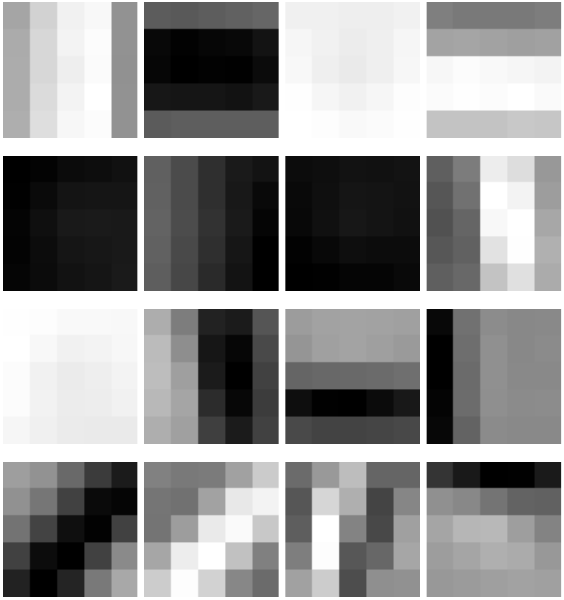}
    \caption{Example learned convolutional dictionary used in the benchmark. The filters are trained offline on Set12 and then kept fixed during inference experiments.}
    \Description{A grid of grayscale convolutional kernels showing an example learned dictionary used for convolutional sparse coding.}
    \label{fig:example_dict}
\end{figure}

Figure~\ref{fig:example_dict} shows an example learned dictionary used in the benchmark.

The benchmark includes both \(3\times3\) and \(5\times5\) convolutional dictionaries with different numbers of kernels, as well as two stride regimes. In particular, stride-1 configurations are evaluated at target sizes \(50\times50\), \(100\times100\), and \(200\times200\), while stride-2 configurations are evaluated at target sizes \(400\times400\) and \(800\times800\), with the latter corresponding to full image size. The sparsity parameter \(\lambda\) is swept over the values \(0.15\), \(0.25\), \(0.5\), \(0.75\), and \(1.5\). Each configuration is repeated over multiple seeds in order to separate systematic effects from run-to-run variation.

In the present study, the analysis is restricted to \(3\times3\) and \(5\times5\) filters. Preliminary experiments with larger filters did not yield sufficiently reliable behavior in our current fixed-point Loihi~2 deployment and hardware mapping, and were therefore excluded from the benchmark. We consequently focus on the convolutional regimes that could be evaluated robustly and reproducibly within the present setup. % wie formulieren??

This design allows us to study convolutional LCA on Loihi~2 over a structured range of operating regimes rather than at a single operating point. The resulting benchmark spans different receptive field sizes, feature-map dimensions, and sparsity levels, thereby exposing how convolutional structure affects quality, latency, power, and energy.

\subsection{Loihi~2 Hardware Platform}
\label{sec:exp_loihi_hardware}

Neuromorphic processing is executed on Intel Loihi~2, a fully programmable digital neuromorphic architecture with 128 neuromorphic cores per chip and support for up to approximately one million neurons and 120 million synapses per chip \cite{intel_labs_loihi_2021}. In our experiments, the deployed network is executed on an single-chip Loihi~2 system. Loihi~2 supports both binary and graded spikes, with graded spike payloads enabling fine-grained synaptic communication. Neuron behavior is programmable through microcode, allowing arithmetic, logical operations, branching, and spike generation to be implemented directly on-chip. Neuromorphic cores communicate via an asynchronous mesh network, while interaction with the host is performed over Ethernet \cite{intel_labs_loihi_2021}.

For the present study, Loihi~2 is used as a neuromorphic inference substrate for convolutional sparse coding. The convolutional LCA dynamics are mapped to recurrent neuron groups with fixed-point membrane state, precomputed feedforward drive, and local inhibitory interactions derived from pairwise dictionary correlations. This hardware setting is particularly relevant because it allows us to study not only whether convolutional LCA can be realized on neuromorphic hardware, but also how its behavior changes across different convolutional regimes.

\subsection{GPU Baseline and Measurement Protocol}
\label{sec:exp_gpu_baseline}

For cross-platform comparison, we additionally evaluate a GPU implementation of the same convolutional LCA inference problem. The baseline experiments are executed on an NVIDIA RTX A6000 GPU utilizing PyTorch 2.0.1 and CUDA 11.7. This GPU implementation uses the same dictionary files, the same Set12 images, the same $\lambda$ sweep, the same stride and target-size configurations, and the same inference budget of 1000 LCA iterations per run. Feedforward drive, lateral interactions, and reconstruction metrics are computed in standard floating-point arithmetic.

To align the comparison with the Loihi~2 benchmark, GPU runs are also averaged across the same five seeds used in the neuromorphic experiments. For each configuration, inference latency is measured as the execution time of the 1000-iteration LCA inference loop, averaged over multiple repeated trials. Crucially, explicit CUDA synchronization (\texttt{torch.cuda.synchronize()}) is enforced before all timer start and stop events to ensure asynchronous kernel execution is fully captured. In the implementation used here, timing repetitions are performed after a warm-up phase where all required tensors and kernels have been constructed, ensuring the reported latency reflects pure sparse-inference execution rather than one-time setup costs.

Dynamic GPU power is estimated relative to an idle baseline. An idle power level is first measured during a separate sampling window. During active execution, power telemetry is sampled repeatedly while the LCA inference loop is executed continuously for a fixed measurement interval. Average active power minus average idle power defines the reported dynamic power. Dynamic energy per inference is then computed as dynamic power multiplied by the average latency of one full 1000-iteration inference call. As in the Loihi~2 case, the reported GPU energy therefore corresponds to a complete inference run rather than to a single LCA iteration.

\subsection{Evaluation Protocol}
\label{sec:exp_protocol}

For each configuration, the Loihi~2 network is executed for a fixed number of LCA iterations. In all reported experiments, a budget of \(1000\) iterations is used. In practice, this budget was sufficient to obtain visually and metrically near-stable solutions across the tested configurations and was therefore adopted as a conservative common stopping horizon for reproducible benchmarking. It should be interpreted as a standardized evaluation budget rather than as an application-optimal iteration count.

Reconstruction quality is evaluated after readout of the final coefficient state, while hardware measurements are obtained from the on-chip execution and power telemetry. We distinguish between algorithmic metrics, such as reconstruction quality and sparsity, and hardware-dependent metrics, such as latency, power, energy, and derived efficiency measures.

A key aspect of the evaluation is that latency and energy are measured with respect to the actual Loihi~2 inference process. In particular, the on-chip execution time is accumulated across all LCA timesteps, yielding a direct measure of neuromorphic inference duration for a single run. Power telemetry is recorded during execution together with a corresponding idle baseline, allowing dynamic power and dynamic energy to be estimated for the Loihi computation itself. This protocol makes it possible to analyze how sparse iterative inference translates into hardware behavior on both Loihi~2 and the conventional GPU baseline while keeping the core convolutional LCA workload aligned across platforms.

\section{Benchmark Metrics}

\subsection{Reconstruction and Sparsity Metrics}
\label{sec:metrics_quality}

Reconstruction quality is quantified primarily by PSNR computed on the dynamically normalized reconstruction. We additionally report SSIM, MSE, and sparsity-related quantities such as the percentage of zero coefficients and the \(\ell_0\) norm of the inferred code. These metrics are treated as hardware-independent in the sense that they describe the inferred representation and reconstruction outcome rather than the execution substrate itself.

\subsection{Latency and Energy Metrics}
\label{sec:metrics_hardware}

Hardware-dependent metrics are defined directly from the Loihi~2 execution and telemetry signals. The metric \texttt{time\_chip\_ms} denotes the accumulated on-chip execution time over all LCA timesteps for a single inference run. In other words, it is obtained by summing the Loihi runtime contribution of each iteration and therefore reflects the neuromorphic processing time of the iterative inference dynamics themselves. The metric \texttt{time\_wall\_ms} denotes the corresponding host-side wall-clock duration for a full inference run.

Dynamic power is estimated by subtracting an idle baseline from the measured active power during execution. Dynamic energy per inference is then computed as the product of dynamic power and inference time for a full run, and is reported in joules per inference. This distinction is important for interpretation. In the present study, Loihi-only latency analyses use \texttt{time\_chip\_ms} in order to characterize the neuromorphic inference dynamics themselves, whereas cross-platform latency comparisons use end-to-end wall-clock latency in order to compare the elapsed runtime of matched Loihi~2 and GPU inference runs. As a result, the reported hardware metrics can be related meaningfully both to sparse activity on-chip and to practical runtime behavior across platforms.

\section{Results}

\subsection{Loihi-Only Parameter Sweep and Tradeoff Structure}
\label{sec:results_loihi_tradeoff}

We first analyze the Loihi~2 results independently of any GPU baseline in order to characterize how convolutional LCA behaves on neuromorphic hardware as a function of the sparsity parameter \(\lambda\). The analysis is based on the parameter sweep described in Section~\ref{sec:exp_dataset}, with reconstruction metrics defined in Section~\ref{sec:metrics_quality} and hardware metrics defined in Section~\ref{sec:metrics_hardware}. This Loihi-only view is important because it establishes the internal structure of the neuromorphic benchmark before any cross-platform interpretation is imposed.

\begin{figure*}[t]
    \centering
    \includegraphics[width=\textwidth]{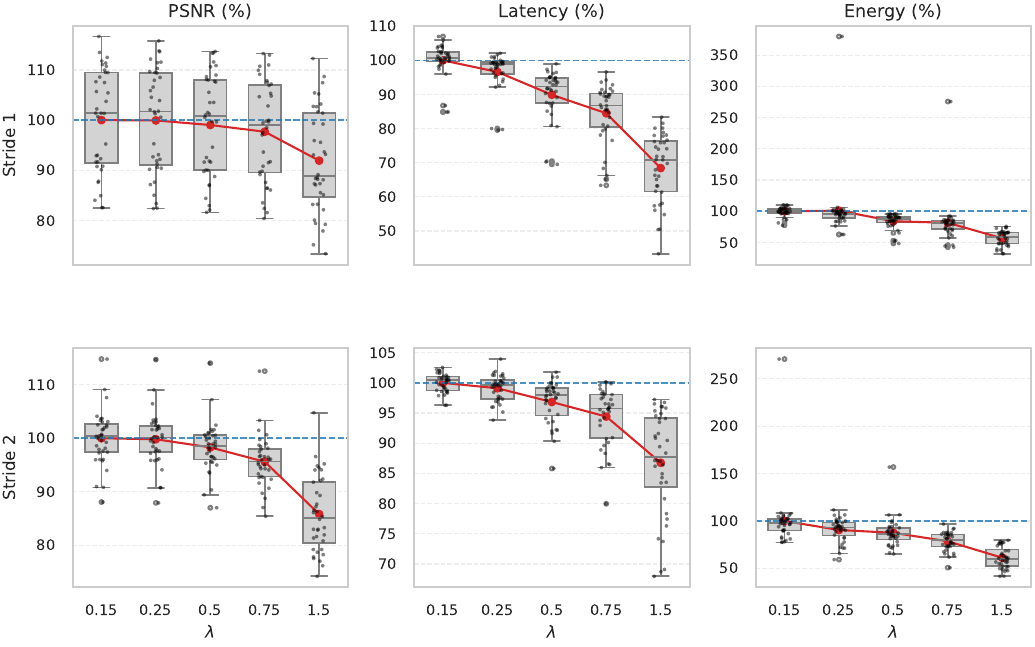}
    \caption{Normalized Loihi-only effect of increasing \(\lambda\) on reconstruction quality, on-chip latency, and dynamic energy for representative stride-1 and stride-2 regimes. For each configuration and image, metrics are first averaged across the five seeds and then normalized to the regime-specific mean at \(\lambda=0.15\), which is set to \(100\%\). Here, latency refers to \texttt{time\_chip\_ms}, i.e., accumulated Loihi on-chip execution time per inference run, and energy refers to dynamic energy per inference in joules before normalization. Boxplots summarize the distribution across Set12 images, while red markers indicate the mean. Increasing \(\lambda\) generally reduces latency and energy, whereas PSNR decreases more moderately.}
    \Description{A full-width figure with normalized boxplots for PSNR, latency, and energy across increasing lambda values, separated into representative stride-1 and stride-2 regimes. Each metric is normalized to its mean value at lambda 0.15 for the corresponding regime.}
    \label{fig:loihi_lambda_sweep}
\end{figure*}

Figure~\ref{fig:loihi_lambda_sweep} summarizes the effect of increasing \(\lambda\) for representative stride-1 and stride-2 regimes. For each configuration and image, metrics are first averaged across the five seeds. The resulting values are then normalized by the regime-specific mean at \(\lambda=0.15\), which is set to \(100\%\). The boxplots therefore show relative changes across Set12 images rather than absolute metric values, making it easier to compare how strongly reconstruction quality, latency, and energy respond to increasing regularization.

The first clear result is that \(\lambda\) acts as a reliable control parameter for sparse inference on Loihi~2. As regularization increases, reconstruction quality tends to decrease, which is consistent with the expected behavior of sparse coding under stronger thresholding. At the same time, increasing \(\lambda\) generally reduces on-chip latency and dynamic energy. The normalized presentation in Figure~\ref{fig:loihi_lambda_sweep} makes clear that these reductions are often more pronounced for latency and energy than for PSNR. A second important observation is that the strength of this effect depends on the convolutional regime. In some settings, increasing \(\lambda\) yields only modest hardware savings, whereas in others the reductions in latency and energy are substantially larger. The figure therefore supports a regime-dependent interpretation: \(\lambda\) is not merely a sparsity parameter, but a practical knob that shifts the operating point of convolutional sparse inference on Loihi~2. This makes the benchmark useful not only for demonstrating feasibility, but also for identifying operating regions in which neuromorphic sparse inference becomes most attractive.

\subsection{Loihi--GPU Comparison}
\label{sec:results_gpu_comparison}

To place the Loihi~2 results in context, we next compare them against a conventional GPU baseline evaluated on the same convolutional LCA inference problems.

\begin{figure}[t]
    \centering
    \includegraphics[width=\columnwidth]{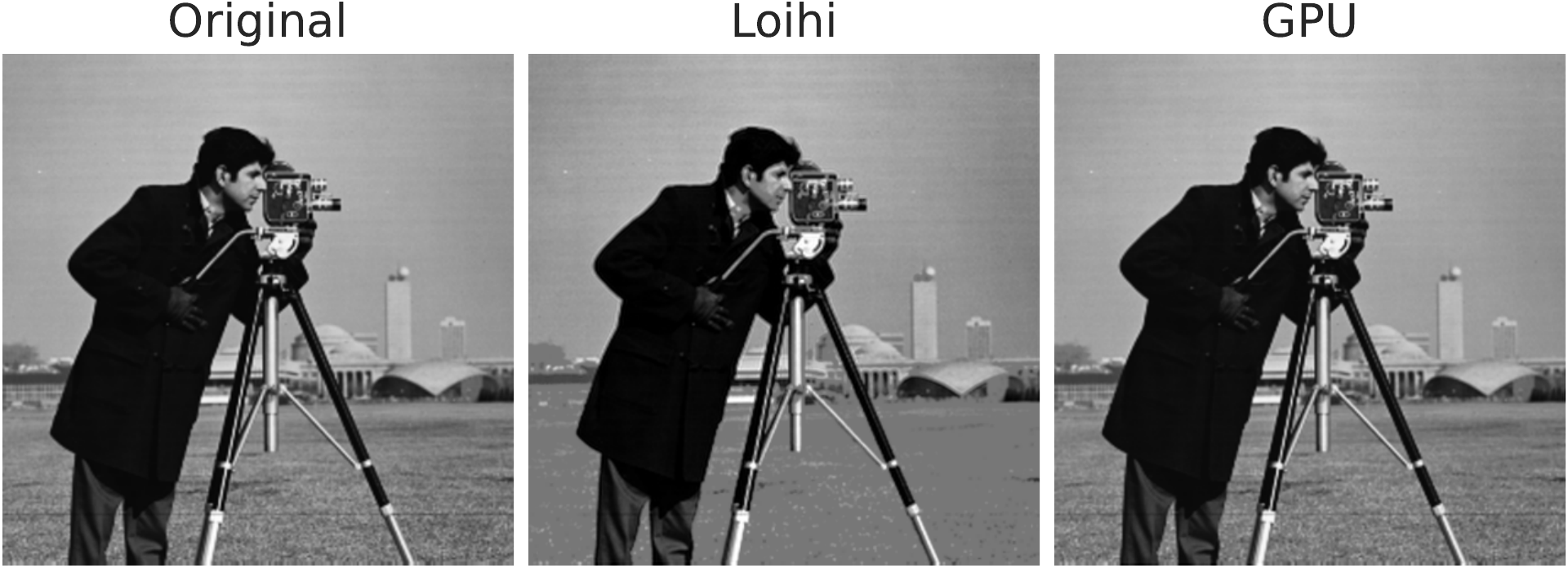}
    \caption{Representative qualitative comparison for one matched Loihi~2 and GPU configuration. Left to right: original Set12 input, Loihi~2 reconstruction, and GPU reconstruction.}
    \Description{Three grayscale panels showing an original Set12 image, a Loihi reconstruction, and a GPU reconstruction for the same convolutional sparse coding configuration.}
    \label{fig:example_reconstruction}
\end{figure}

\begin{figure*}[t]
    \centering
    \includegraphics[width=\textwidth]{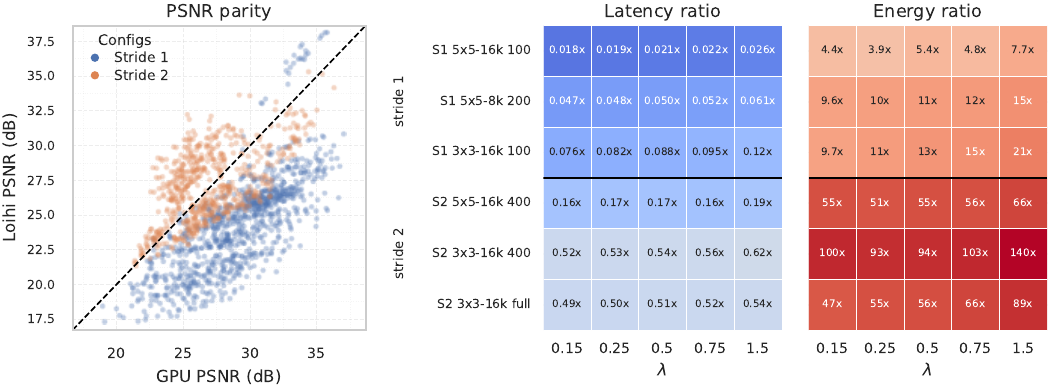}
    \caption{Comparison of matched Loihi~2 and GPU runs over the same convolutional LCA configurations. Left: parity plot for PSNR after averaging across seeds. Middle and right: wall-clock latency and dynamic-energy ratios (GPU / Loihi) for representative regimes across \(\lambda\). Latency ratios are computed from full-run \texttt{time\_wall\_ms}, while energy ratios use dynamic energy per inference in joules. Values above one indicate an advantage for Loihi~2, while values below one indicate an advantage for the GPU.}
    \label{fig:loihi_gpu_comparison}
\end{figure*}

Figure~\ref{fig:example_reconstruction} provides a representative qualitative example, while Figure~\ref{fig:loihi_gpu_comparison} summarizes the quantitative comparison over the full benchmark.

Several patterns are visible in Figure~\ref{fig:loihi_gpu_comparison}. First, the PSNR parity plot shows that cross-platform reconstruction quality is regime-dependent rather than uniformly ordered across all matched runs. Many configurations lie below the parity diagonal, but the deviations are not uniform across stride and size settings, and the apparent separation becomes smaller in higher-\(\lambda\) regimes, where both platforms operate at sparser solutions.

Second, the latency ratios are strongly regime-dependent. In the stride-1 settings, the GPU is consistently faster, often by a large margin. In the stride-2 settings, the gap is reduced, and the relative latency difference depends more strongly on the particular operating regime than on \(\lambda\) alone.

Third, the energy ratios show a consistent advantage for Loihi~2 across the representative regimes considered here. Although this advantage varies in magnitude, Loihi~2 requires substantially less dynamic energy per full inference run than the GPU baseline throughout the comparison.

\begin{table*}[t]
\centering
\caption{Absolute cross-platform hardware comparison for selected representative convolutional LCA regimes. Metrics are averaged across seeds first and then across Set12 images. Latency is reported as full-run wall-clock latency in milliseconds, and energy is reported as dynamic energy per inference in millijoules.}
\label{tab:absolute_hardware_comparison_representative}
\small
\setlength{\tabcolsep}{4pt}
\begin{tabular}{llrrrr}
\toprule
Regime & \(\lambda\) & Latency Loihi (ms) & Latency GPU (ms) & Energy Loihi (mJ) & Energy GPU (mJ) \\
\midrule
S1 5x5-16k 100  & 0.15 & 5549.9 & 101.1 & 3464.86 & 15093.26 \\
S1 5x5-16k 100  & 1.50 & 3862.2 & 101.5 & 2004.33 & 15513.39 \\
S1 3x3-16k 100  & 0.15 & 1376.7 & 105.2 & 1072.69 & 10457.52 \\
S1 3x3-16k 100  & 1.50 &  847.9 & 104.2 &  503.33 & 10651.83 \\
S2 5x5-16k 400  & 0.15 &  515.7 &  84.6 &  304.24 & 16598.34 \\
S2 5x5-16k 400  & 1.50 &  461.5 &  85.8 &  225.64 & 14829.62 \\
S2 3x3-16k full & 0.15 &  204.8 &  99.7 &  496.98 & 23130.22 \\
S2 3x3-16k full & 1.50 &  191.0 & 102.6 &  299.27 & 26491.64 \\
\bottomrule
\end{tabular}
\end{table*}

Table~\ref{tab:absolute_hardware_comparison_representative} provides an absolute-value anchor for the latency and energy ratios in Figure~\ref{fig:loihi_gpu_comparison}. The stride-1 regimes show a large latency advantage for the GPU, whereas the stride-2 regimes reduce this gap considerably. Across all representative regimes, however, Loihi~2 requires substantially less dynamic energy per inference. The table therefore complements the ratio plots by showing that the observed cross-platform hardware differences correspond to substantial differences in absolute operating cost.

Taken together, these results show that the relative behavior of the two platforms is not uniform: the GPU provides substantially lower latency, whereas Loihi~2 provides a strong dynamic-energy advantage. Reconstruction quality exhibits a more regime-dependent cross-platform pattern and is therefore best interpreted through the parity analysis rather than as a single global ordering. The magnitude of these differences depends strongly on the convolutional regime and operating point.

\section{Discussion}

The results position convolutional LCA on Loihi~2 as a regime-dependent sparse-inference substrate rather than a direct replacement for conventional GPU execution. Across the matched benchmark, the GPU provides lower latency, while Loihi~2 offers a substantial advantage in dynamic energy per full inference run. Reconstruction quality shows a more regime-dependent cross-platform pattern and should therefore be interpreted with greater care than the hardware metrics alone. 

The Loihi-only sweeps show that this design space already has substantial internal structure before any cross-platform comparison is imposed. In particular, the effect of \(\lambda\) is not merely to change sparsity in the abstract; it reshapes the practical tradeoff between reconstruction fidelity and hardware cost in a way that depends on stride, filter size, and target regime. The GPU comparison then shows that these same regimes map differently onto conventional and neuromorphic hardware. This motivates interpreting convolutional LCA not as a single benchmark point, but as a family of operating points whose usefulness depends on the system objective.

A plausible interpretation of the observed cross-platform quality differences is that the fixed-point Loihi~2 realization is most challenged in denser operating regimes, where many small coefficient interactions must be represented and accumulated over long recurrent dynamics. As \(\lambda\) increases and the inferred representations become sparser, these precision-related effects may become less detrimental, which is broadly consistent with the reduced separation seen in some higher-\(\lambda\) settings. This interpretation remains qualitative, but it aligns with the observed trend across the matched runs.

From an application perspective, the present benchmark is most favorable to scenarios in which iterative sparse inference is valuable but absolute latency is not the only objective. Examples include energy-constrained sensing pipelines, always-on front-end processing, or settings in which sparse inference can be warm-started from previous states rather than solved from scratch each time. In such cases, the energy advantage of Loihi~2 may be more important than the fact that a conventional workstation GPU remains faster in absolute terms.

At the same time, the fixed budget of \(1000\) LCA iterations used here should be interpreted as a conservative benchmarking horizon rather than as a claim that all practical deployments require the same number of iterations. In many applications, useful solutions may be obtained with fewer steps, especially when warm starts, temporal continuity, or task-specific stopping criteria are available. The present benchmark does not evaluate such strategies, but it helps identify the operating regimes in which they may be most promising.

The present results should therefore not be overgeneralized. The GPU baseline used here is a strong conventional reference that was readily available, not a hardware-class-matched neuromorphic counterpart. Its role is to contextualize the Loihi~2 results against a widely used conventional compute substrate. Under that interpretation, the comparison highlights the limits of the current Loihi~2 mapping as clearly as its strengths. The contribution of the benchmark is not to show universal neuromorphic superiority, but to identify where convolutional sparse inference on Loihi~2 is feasible, where it is advantageous, and where conventional hardware remains preferable.

\section{Limitations}

Several limitations should be kept in mind when interpreting the present results. First, the benchmark is restricted to a relatively narrow family of convolutional sparse-coding problems. In particular, we consider only offline-trained \(3\times3\) and \(5\times5\) dictionaries evaluated on Set12 images under a one-layer recurrent convolutional LCA formulation. Larger filters were explored during development but did not yield sufficiently robust behavior in the current fixed-point Loihi~2 deployment, and were therefore excluded from the reported study. The present results should consequently be understood as a controlled benchmark of convolutional sparse inference rather than as a comprehensive evaluation of convolutional sparse coding more broadly.

Second, the benchmark comparison is intentionally asymmetric in hardware class. The GPU baseline is a workstation-class device that is expected to provide very high throughput, whereas Loihi~2 is designed around different architectural objectives, especially sparse event-driven computation and energy-efficient inference. The comparison is still informative because both platforms solve the same matched inference problems, but it should not be interpreted as a surprising head-to-head contest in which equal absolute latency would normally be expected. Rather, it clarifies the tradeoff between reconstruction quality, runtime, and dynamic energy across two very different computing substrates.

Third, the Loihi~2 implementation is subject to fixed-point quantization and hardware-specific mapping constraints. These constraints likely contribute to the systematic reconstruction-quality gap relative to the GPU baseline, especially in lower-\(\lambda\) regimes where coefficient activity is denser and small numerical effects can accumulate more strongly. The study therefore compares a realizable fixed-point neuromorphic deployment against a high-performance floating-point GPU implementation, not two numerically equivalent realizations. For this reason, cross-platform differences in reconstruction quality should be interpreted as benchmark outcomes of the specific implemented pipelines rather than as a pure precision-only comparison.

Fourth, the cross-platform comparison uses a fixed budget of \(1000\) LCA iterations per inference run. This choice provides a controlled and reproducible benchmark across all regimes, but it is not necessarily representative of practical deployment. In application settings such as image denoising or temporally correlated inputs, useful solutions may often be obtained with substantially fewer iterations, especially when warm starts \cite{kasenbacher2025warp} or adaptive convergence criteria are used. The present study does not evaluate such strategies, and therefore likely reflects an upper-end inference cost rather than an application-optimized operating point.

Fifth, the cross-platform comparison should be interpreted as a matched benchmark rather than as an exact like-for-like physical equivalence. The same convolutional LCA inference problem is evaluated across platforms using aligned configurations, but latency, power, and energy are necessarily measured through platform-specific instrumentation. In both cases, energy is reported per full inference run rather than per single iteration, and dynamic power is estimated relative to an idle baseline. These choices make the comparison meaningful for benchmarking, but they do not eliminate all differences in measurement semantics, host interaction, or system overhead.

Finally, the paper focuses exclusively on inference. Dictionary learning is performed offline, no on-chip adaptation is considered, and no end-to-end application pipeline is evaluated. The results therefore clarify the behavior of convolutional sparse inference on Loihi~2, but they do not yet establish the full system-level usefulness of such models in broader neuromorphic applications.

\section{Conclusion}

We presented a Loihi~2 implementation of convolutional sparse coding via the Locally Competitive Algorithm and studied it as a benchmark for structured sparse inference on neuromorphic hardware. The implementation realizes a one-layer recurrent convolutional LCA with precomputed feedforward drive, locally structured inhibitory interactions, and fixed-point on-chip dynamics. To the best of our knowledge, this is the first benchmark study of convolutional LCA on Loihi~2.

The main result is that convolutional LCA on Loihi~2 is feasible, but strongly regime-dependent. The Loihi-only analysis shows that the quality--efficiency tradeoff depends systematically on \(\lambda\), stride, filter size, and target regime rather than collapsing to a single operating point. The cross-platform comparison adds that a conventional workstation GPU generally provides lower latency, whereas Loihi~2 offers a substantial dynamic-energy advantage across the representative regimes considered here. Reconstruction quality shows a more regime-dependent cross-platform pattern and does not reduce to a single uniform ordering.

The practical implication is therefore not that Loihi~2 uniformly outperforms conventional hardware, but that it occupies a different part of the design space. GPU execution remains preferable when absolute latency and reconstruction quality are the dominant objectives, whereas Loihi~2 becomes attractive when iterative sparse inference must be carried out under tight dynamic-energy constraints. More broadly, the results suggest that convolutional LCA is a useful neuromorphic benchmark precisely because it exposes a structured family of tradeoffs whose shape depends on both operating regime and hardware substrate.

\begin{acks}
This work was carried out in the publicly funded NAOMI4Radar project, supported by the German Federal Ministry for Economic Affairs and Climate Action (BMWK) under Grant No. 19A24001A. The authors would also like to thank Sumedh R. Risbud and Philipp Plank from Intel Labs for their helpful discussions.
\end{acks}

\bibliographystyle{ACM-Reference-Format}
\bibliography{references}

\end{document}